\documentclass{article}

\PassOptionsToPackage{numbers,compress}{natbib}

\usepackage[preprint]{neurips_2026}

\usepackage[utf8]{inputenc}
\usepackage[T1]{fontenc}
\usepackage{hyperref}
\usepackage{url}
\usepackage{booktabs}
\usepackage{amsfonts}
\usepackage{nicefrac}
\usepackage{microtype}
\usepackage{xcolor}

\usepackage{graphicx}
\usepackage{subcaption}
\usepackage{amsmath}
\usepackage{amssymb}
\usepackage{mathtools}
\usepackage{amsthm}
\usepackage{multirow}
\usepackage{threeparttable}
\usepackage{pifont}
\usepackage{wrapfig}

\usepackage[capitalize,noabbrev]{cleveref}

\hypersetup{
  pdftitle={GeoUniPR: A Geometry-Consistent Unified Framework for Cross-Modal Place Recognition},
  pdfauthor={Wonbong Kim, Jiatong Xiao, Rui Li, Xufei Wang, Qiwen Gu, Junqiao Zhao, Chen Ye, Guang Chen}
}

\title{GeoUniPR: A Geometry-Consistent Unified Framework for Cross-Modal Place Recognition}

\author{%
  \normalfont
  Wonbong Kim$^{1}$,
  Jiatong Xiao$^{1}$,
  Rui Li$^{1}$,
  Xufei Wang$^{3}$
  \\[1mm]
  Qiwen Gu$^{1}$,
  Junqiao Zhao$^{1,2,*}$,
  Chen Ye$^{1}$,
  Guang Chen$^{1,4}$
  \\[2mm]
  \small
  $^{1}$School of Computer Science and Technology, Tongji University, Shanghai, China \\
  $^{2}$MOE Key Lab of Embedded System and Service Computing, \\
  Tongji University, Shanghai, China \\
  $^{3}$Shanghai Research Institute for Intelligent Autonomous System, \\
  Tongji University, Shanghai, China \\
  $^{4}$Shanghai Innovation Institute, Shanghai, China
  \\[1mm]
  \texttt{\{2493052,2354218,2351610,tjwangxufei,2432178,}%
  \\
  \texttt{zhaojunqiao,yechen,guangchen\}@tongji.edu.cn}
}

\begin{document}

\maketitle

\begingroup
\renewcommand{\thefootnote}{*}
\footnotetext{Corresponding author.}
\endgroup

\begin{abstract}
Cross-modal place recognition (CMPR) aims to identify the same location across heterogeneous sensing modalities, such as vision and LiDAR.
Existing methods commonly bridge the modality gap using complex alignment modules, multi-stage training, or full fine-tuning of pretrained backbones.
In this work, we revisit CMPR from the perspective of geometric consistency and propose GeoUniPR, a unified and concise geometry-consistent framework.
GeoUniPR reduces cross-modal discrepancy at the representation level by projecting LiDAR point clouds into the camera perspective to construct Geometry-Consistent depth image views (DIV), which establish direct RGB--LiDAR correspondence.
We further augment DIV with native LiDAR cues, including intensity and surface-normal information, yielding a multi-channel geometric representation that improves structural consistency.
Based on this representation, GeoUniPR learns a unified embedding space using two modality-specific ViT-based encoders with identical architectures, trained through parameter-efficient adaptation without auxiliary alignment modules, multi-stage training, or full backbone fine-tuning.
In addition, we introduce Spatially-Consistent InfoNCE (SC-InfoNCE), a CMPR-specific contrastive objective that suppresses distance-induced false negatives under spatial continuity.
Extensive experiments on KITTI and KITTI-360 demonstrate that GeoUniPR achieves state-of-the-art (SOTA) performance in both same-modal and cross-modal place recognition, with strong cross-dataset generalization.
\end{abstract}

\section{Introduction}
Place recognition \cite{Yin25GPRSurvey} is a fundamental capability in autonomous systems, supporting applications such as navigation \cite{Cadena16SLAMSurvey}, loop closure \cite{Newman02ExploreReturn}, map reuse \cite{Ratz20OneShot}, and global localization \cite{Sarlin19CoarseToFine}. 
Traditional place recognition methods are typically studied under a single-modality setting \cite{Barros21PRSurvey}, where both the query and the reference database are captured using the same sensing modality.

In real-world autonomous systems, however, sensing configurations are often heterogeneous.
Differences between the sensing modalities of queries and reference maps naturally give rise to Cross-Modal Place Recognition (CMPR), which aims to localize a query from one modality (e.g., vision or LiDAR) by retrieving its corresponding place from a database constructed using another modality.
By enabling retrieval across vision and LiDAR, CMPR unifies both same-modal (2D$\rightarrow$2D, 3D$\rightarrow$3D) and cross-modal (2D$\rightarrow$3D, 3D$\rightarrow$2D) localization, allowing autonomous systems to reuse existing maps and remain robust under changing sensing conditions or sensor failures.

The core challenge of CMPR lies in the substantial heterogeneity between sparse geometric 3D point clouds and dense texture-rich 2D images.
To bridge this modality gap, many recent approaches \cite{Xia24UniLoc,Jiao25InsCMPR,Lee23LiDARCameraLoopConstraints} rely on increasingly complex architectures, typically by introducing auxiliary alignment modules \cite{Xia24UniLoc,Jiao25InsCMPR}, adopting multi-stage optimization pipelines \cite{Lee23LiDARCameraLoopConstraints}, or fully fine-tuning pretrained backbones \cite{Puligilla24LIPLoc,Li25VXP}.
These methods largely follow a feature-level alignment paradigm: RGB and LiDAR inputs are first encoded into modality-specific feature spaces, after which additional modules or optimization strategies are introduced to reduce the discrepancy between the learned embeddings.
However, this paradigm leaves the underlying geometric inconsistency unresolved.
When the two modalities are encoded under incompatible projection geometries, subsequent feature-level alignment must simultaneously compensate for perspective mismatch, sensor-specific distortions, and modality appearance gaps, making the learned correspondence unnecessarily indirect and fragile.

Meanwhile, CMPR pipelines also attempt to reduce modality discrepancy through geometric projection.
Common representations include Range Image View (RIV) \cite{Puligilla24LIPLoc}, Depth Image View (DIV) \cite{Jiao25InsCMPR}, and Bird’s-Eye-View (BEV) alignment \cite{Zheng23I2PRec}.
These parameterizations expose a fundamental trade-off between geometric correspondence and information integrity.
RIV adopts spherical projection and is therefore geometrically inconsistent with the camera perspective, preventing direct pixel-wise RGB--LiDAR correspondence and typically requiring explicit overlap handling (e.g., horizontal FOV alignment).
DIV naturally preserves pixel-aligned correspondence under the camera model, but becomes intrinsically sparse after projection.
BEV unifies both modalities on the ground plane, yet is mainly applicable to driving scenarios and requires an additional lift-and-project pipeline.

\begin{wrapfigure}{r}{0.40\textwidth}
  \centering
  \includegraphics[width=\linewidth]{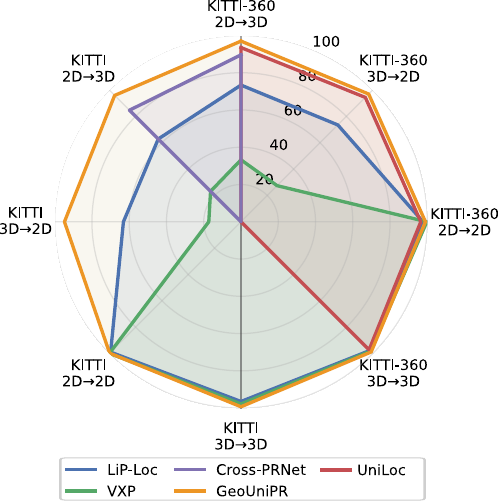}
  \caption{
      GeoUniPR achieves SOTA performance on KITTI and KITTI-360.
      Note that Cross-PRNet reports only 2D$\to$3D results, and UniLoc reports results only on KITTI-360, so the remaining entries are unavailable.
  }
  \label{fig:radar}
\end{wrapfigure}

These observations motivate a different perspective: instead of compensating for modality discrepancy after feature extraction, CMPR should first establish geometry-consistent representations before learning modality-invariant descriptors.
Among existing geometric parameterizations, camera-perspective DIV provides the most direct way to preserve pixel-wise RGB--LiDAR correspondence under a shared projection geometry.
Based on this principle, we propose GeoUniPR, a geometry-consistent representation learning framework for CMPR.
Rather than relying on explicit feature alignment modules, GeoUniPR reduces modality discrepancy before feature extraction by projecting LiDAR scans into camera-aligned multi-channel DIVs.

Specifically, LiDAR point clouds are projected into the camera perspective to construct multi-channel DIVs jointly encoding depth, intensity, and surface-normal cues.
The resulting DIVs are processed using modality-specific ViT-based encoders \cite{Caron21DINO,Oquab23DINOv2,Simeoni25DINOv3} with identical architectures and parameter-efficient adaptation \cite{Lu24CricAVPR,Lu25SelaVPRpp}.
RGB and LiDAR embeddings are then aligned in a unified representation space through contrastive learning \cite{Radford21CLIP}.
To further improve cross-modal optimization, we introduce Spatially-Consistent InfoNCE (SC-InfoNCE), which relaxes the strict negative assumption in conventional InfoNCE by suppressing spatially adjacent false negatives under trajectory continuity.

The main contributions of this work are summarized as follows:
\begin{itemize}
    \item We propose GeoUniPR, a geometry-consistent representation learning framework for CMPR that shifts cross-modal alignment from post-hoc feature matching to geometry-consistent representations. GeoUniPR constructs camera-perspective multi-channel DIVs to establish direct RGB--LiDAR correspondence without auxiliary alignment modules, multi-stage training, or full backbone fine-tuning.
    
    \item We introduce Spatially-Consistent InfoNCE (SC-InfoNCE), which mitigates distance-induced false negatives by suppressing or reweighting spatially adjacent negatives, thereby improving contrastive learning under spatial continuity.
    
    \item We conduct extensive experiments on KITTI and KITTI-360, demonstrating that GeoUniPR achieves state-of-the-art performance across both same-modal and cross-modal retrieval settings, while exhibiting strong cross-dataset generalization.
\end{itemize}

\section{Related Work}

\subsection{Cross-Modal Place Recognition}
\label{sec:related_cmpr}

The key challenge of CMPR is to learn descriptors that are simultaneously modality-aligned and place-discriminative despite the heterogeneous 2D appearance and 3D geometry.
Cattaneo et al.~\cite{Cattaneo20SharedEmbedding} pioneered the application of knowledge distillation to CMPR, employing a teacher--student framework to train a 3D network that mimics the feature space of a 2D image network using triplet loss.
Later, LC$^2$ \cite{Lee23LiDARCameraLoopConstraints} improves cross-modal matching by introducing a shared intermediate representation and optimizing it with a staged training procedure.
LiP-Loc \cite{Puligilla24LIPLoc} pioneers the use of pretrained ViT backbones in CMPR, enabling direct cross-modal descriptor learning via contrastive loss.
VXP~\cite{Li25VXP} further strengthens cross-modal coupling through a teacher-guided framework that aligns local descriptors between modalities via voxel-to-pixel correspondence.
UniLoc \cite{Xia24UniLoc} extends CMPR to a universal setting by aligning natural language, images, and point clouds within a shared embedding space using hierarchical instance-level and scene-level matching.
Most recently, Cross-PRNet~\cite{Meng25ContrastiveIF} introduces a Transformer--Mamba based design to enhance contextual modeling for learning discriminative cross-modal place descriptors.
Overall, prior CMPR methods often rely on auxiliary modules or staged optimization, and some further adopt full backbone fine-tuning to obtain strong cross-modal retrieval performance.

\subsection{Geometric Consistency in Cross-Modal Place Recognition}
\label{sec:related_geom}

A common line of work in CMPR seeks to induce geometrically meaningful 2D--3D associations by reducing the representational gap between images and point clouds. To this end, projection-based representations are widely adopted, mapping one or both modalities into structured intermediate views that facilitate cross-modal alignment and retrieval.

\textbf{RIV-based consistency.}
A common design is to project LiDAR scans into range image views (RIV) for dense matching.
LiP-Loc \cite{Puligilla24LIPLoc} follows this paradigm and applies Field of View (FOV)-aligned cropping to match the camera view, reducing mismatch within the shared FOV.
LC$^2$ \cite{Lee23LiDARCameraLoopConstraints} further aligns modalities in a shared 2.5D space by lifting images to depth via depth estimation and projecting LiDAR into RIV, enabling correspondence learning under staged optimization.

\textbf{DIV-based consistency.}
Another line of work projects LiDAR onto the image plane as depth image view (DIV), aligning 3D geometry with the camera perspective.
VXP \cite{Li25VXP} voxelizes point clouds with sparse convolution and projects voxel features onto the pixel plane to enforce voxel-to-pixel coupling guided by an image teacher.
InsCMPR \cite{Jiao25InsCMPR} similarly obtains image-plane DIV via camera projection and further applies depth densification to better match the dense image grid.

\textbf{BEV-based consistency.}
Beyond perspective views, I2P-Rec \cite{Zheng23I2PRec} lifts images to 3D through depth estimation, reconstructs pseudo point clouds, and projects both modalities into BEV space for retrieval against point-cloud maps.


In contrast to prior works that use geometric projections mainly as preprocessing or auxiliary alignment signals, GeoUniPR treats projection geometry as the central representation-learning principle. The key distinction is that we seek to minimize cross-modal discrepancy before feature extraction, rather than relying on downstream feature interaction modules to repair modality mismatch after encoding.
Besides, existing CMPR pipelines primarily rely on the geometric channel (range or depth) and, to the best of our knowledge, have not incorporated LiDAR-native cues such as intensity and surface normals.

\section{Preliminaries}
\label{sec:preliminaries}

\subsection{Problem Formulation}
Let $\mathcal{D}_I = \{x_i^I\}_{i=1}^N$ and $\mathcal{D}_P = \{x_i^P\}_{i=1}^N$ denote paired RGB--LiDAR observations, where each pair $(x_i^I, x_i^P)$ is captured at the same place.
The goal is to learn two encoders $f_I: \mathcal{X}_I \to \mathbb{R}^d$ and $f_P: \mathcal{X}_P \to \mathbb{R}^d$ that map heterogeneous inputs into a shared embedding space.

During inference, we perform cross-modal place recognition via similarity-based retrieval in the shared embedding space.
Given a query from one modality, we rank samples from the other-modality database by embedding similarity and return the top-$K$ candidates (for both $I\!\to\!P$ and $P\!\to\!I$).
A prediction is considered correct if the spatial distance between the query location and the retrieved match is within a tolerance threshold $\delta$.

\subsection{Cross-Modal Contrastive Learning}
\label{sec:prelim_infonce}

Given a batch of $N$ paired RGB--LiDAR samples $\{(x_i^{I}, x_i^{P})\}_{i=1}^{N}$, we encode them into $\ell_2$-normalized embeddings
$\mathbf{z}_i^{I}=f_I(x_i^{I})/\|f_I(x_i^{I})\|_2$ and
$\mathbf{z}_i^{P}=f_P(x_i^{P})/\|f_P(x_i^{P})\|_2$.
The cross-modal matching logit is defined as
\begin{equation}
    s_{ij}=\frac{(\mathbf{z}_i^{I})^\top \mathbf{z}_j^{P}}{\tau},
    \label{eq:logits}
\end{equation}
where $\tau$ is the temperature.

Standard InfoNCE optimizes RGB-to-LiDAR retrieval by aligning each positive pair $(\mathbf{z}_i^{I},\mathbf{z}_i^{P})$ and treating other in-batch samples as negatives:
$\mathcal{L}_{I\to P}^{\mathrm{InfoNCE}}
= -\frac{1}{N}\sum_{i=1}^{N}
\log \frac{\exp(s_{ii})}{\sum_{j=1}^{N}\exp(s_{ij})}$.
Following CLIP~\cite{Radford21CLIP}, we use a symmetric objective by also optimizing the reverse direction,
$\mathcal{L}_{P\to I}^{\mathrm{InfoNCE}}
= -\frac{1}{N}\sum_{i=1}^{N}
\log \frac{\exp(s_{ii})}{\sum_{j=1}^{N}\exp(s_{ji})}$.
The final loss is
$\mathcal{L}
= \frac{1}{2}(\mathcal{L}_{I\to P}^{\mathrm{InfoNCE}}
+\mathcal{L}_{P\to I}^{\mathrm{InfoNCE}})$

\section{Methodology}
\begin{figure*}[t]
  \centering
  \includegraphics[width=0.96\textwidth]{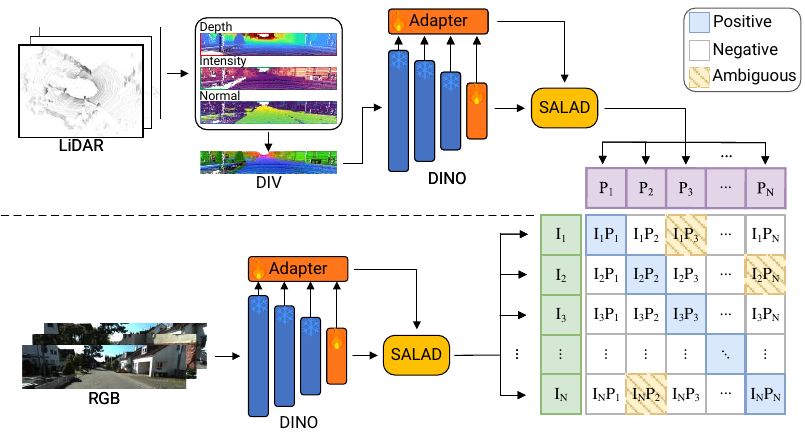}
  \caption{
      Overview of GeoUniPR framework.
      LiDAR scans are projected and densified into a DIV containing depth, intensity, and normal ratio channels to establish pixel-wise geometric alignment with RGB images.
      Both modalities are encoded by modality-specific DINO ViT backbones \cite{Caron21DINO} with lightweight adapters and aggregated via SALAD \cite{Izquierdo24SALAD}.
      To improve training, we employ a SC-InfoNCE loss that explicitly handles spatially ambiguous regions by suppressing or reweighting nearby non-matching pairs, effectively mitigating distance-induced false negatives.
  }
  \label{fig:method}
\end{figure*}

\subsection{Overview}
\label{sec:overview}

\cref{fig:method} presents an overview of the proposed GeoUniPR framework.
GeoUniPR is built upon a representation-first view of CMPR.
Instead of encoding RGB and LiDAR under incompatible geometric parameterizations and compensating for the resulting discrepancy afterward, we first transform LiDAR into a camera-perspective representation that shares the same projection geometry as RGB images.
This enables the two modalities to be processed by architecturally identical encoders and aligned through a simple contrastive objective.

Specifically, we instantiate two modality-specific encoders with identical architectures but separate parameters for RGB and DIV inputs.
As the backbone, we adopt a pre-trained DINO ViT encoder \cite{Caron21DINO,Oquab23DINOv2,Simeoni25DINOv3} and perform parameter-efficient fine-tuning (PEFT) following \cite{Lu25SelaVPRpp}.
In particular, lightweight MultiConv adapters \cite{Lu24CricAVPR} are attached in parallel to selected frozen transformer blocks at fixed intervals, enabling the model to adapt intermediate patch representations to domain-specific geometric patterns while preserving the semantic priors of the pre-trained backbone.
For global descriptor aggregation, we employ SALAD \cite{Izquierdo24SALAD}, which maps adapter-refined local patch features into learnable clusters through optimal transport (Sinkhorn iterations) and fuses them with the global \texttt{[CLS]} token produced by the ViT backbone, yielding compact and discriminative representations for cross-modal retrieval.

Built upon this unified and efficient framework, we further introduce geometry-consistent designs tailored to heterogeneous RGB--LiDAR inputs.
At the representation level, LiDAR point clouds are projected into a normal-augmented multi-channel DIV (\cref{sec:div_generation}), establishing strict pixel-wise correspondence with RGB observations.
At the optimization level, we introduce SC-InfoNCE (\cref{sec:sc_infonce}), a spatially-consistent contrastive objective that suppresses false negatives arising from spatially continuous trajectories, thereby stabilizing cross-modal embedding learning.

\subsection{Geometry-Consistent Depth Image View}
\label{sec:div_generation}

We adopt DIV as the intermediate representation to establish a geometry-consistent bridge between RGB images and LiDAR point clouds.
By projecting 3D LiDAR points onto the camera plane, we naturally resolve the horizontal FOV discrepancy and obtain a perspective-consistent, pixel-aligned signal that directly corresponds to the RGB image.

\paragraph{Perspective Projection.}
Given a point cloud $\mathcal{P}$, we project each point onto the image plane using the camera intrinsic matrix $\mathbf{K}$ and the LiDAR-to-camera extrinsic transformation $\mathbf{T}_{v\to c}$.
For a point $\mathbf{p}_i$, its image coordinate is computed as
$z_i \tilde{\mathbf{u}}_i = \mathbf{K}\mathbf{T}_{v\to c}\tilde{\mathbf{p}}_i$.
Since the mapping from $3\mathrm{D}$ to $2\mathrm{D}$ is non-injective, multiple points may project onto the same pixel.
We resolve such collisions using z-buffering, retaining only the closest point for each pixel.

\paragraph{Multi-channel DIV Construction and Densification.}
We construct a camera-aligned multi-channel map
$\mathcal{I}_{\mathrm{div}} \in \mathbb{R}^{H \times W \times 3}$,
which jointly encodes complementary geometric cues:
(1) metric depth, serving as the primary geometric signal after perspective projection;
(2) intensity, providing an additional sensor-dependent cue complementary to depth;
and (3) normal ratio, a local surface-structure descriptor computed from the eigenvalue ratio of the neighborhood covariance matrix \cite{Jung25ImLPR}.

Because perspective projection produces sparse observations on the image plane, we adopt the training-free IP-Basic densification method \cite{Ku18DepthCompletionCPU} to propagate valid measurements into unobserved regions, resulting in a dense pixel-aligned representation.
Finally, we crop the upper non-overlapping region to account for the vertical FOV discrepancy between LiDAR and camera views.

\subsection{Spatially-Consistent Contrastive Learning}
\label{sec:sc_infonce}

\paragraph{Unified Spatially-Consistent Weighting.}

Standard InfoNCE assumes that for any anchor sample $i$, all distinct samples $j \neq i$ correspond to true negatives.
However, place recognition data are inherently spatially continuous.
Samples captured at nearby locations often share significant geometric and semantic overlap.
Treating such spatially adjacent samples as hard negatives forces the model to separate highly similar observations, leading to the false-negative problem.

To suppress distance-induced false negatives, we propose Spatially-Consistent Contrastive Learning (SC-InfoNCE), which introduces distance-dependent reweighting into the InfoNCE denominator.
Let $\mathbf{g}_i$ denote the metric location of sample $i$ and $d_{ij}=\|\mathbf{g}_i-\mathbf{g}_j\|_2$ denote the pairwise distance.
We define a weighting function $\omega:\mathbb{R}_{\ge 0}\rightarrow[0,1]$ and formulate the RGB$\!\to$LiDAR loss as:
\begin{equation}
\mathcal{L}_{I\to P}^{\mathrm{SC}}(\omega)
= -\frac{1}{N}\sum_{i=1}^{N}
\log
\frac{\exp(s_{ii})}{
\sum_{j=1}^{N}\omega(d_{ij})\,\exp(s_{ij})
},
\label{eq:sc_infonce}
\end{equation}
where $s_{ij}$ is defined in \cref{eq:logits}.
We enforce $\omega(d_{ii})=1$ to preserve positive pairs.
In practice, we employ the symmetric bidirectional formulation by averaging \cref{eq:sc_infonce} with its LiDAR$\!\to$RGB counterpart, and omit the reverse-direction expression for brevity.

Within this unified framework, we consider two instantiations of $\omega(\cdot)$: hard masking and hybrid reweighting.

\paragraph{Spatially-Consistent Hard Masking (SC-Hard).}
SC-Hard suppresses spatially ambiguous pairs through an ambiguous-region radius $R_{\mathrm{n}}$.
For each anchor $i$, all off-diagonal pairs satisfying $0 < d_{ij} \le R_{\mathrm{n}}$ are removed from the contrastive denominator, thereby preventing false-negative penalization.
Equivalently, this can be implemented through the binary distance mask:
\begin{equation}
\omega_{\mathrm{hard}}(d_{ij})=
\begin{cases}
1, & d_{ij}=0,\\
0, & 0<d_{ij}\le R_{\mathrm{n}},\\
1, & d_{ij}>R_{\mathrm{n}}.
\end{cases}
\label{eq:omega_hard}
\end{equation}

\paragraph{Spatially-Consistent Hybrid Reweighting (SC-Hybrid).}
While SC-Hard is simple and effective, it treats all ambiguous pairs equally.
SC-Hybrid instead provides a smooth relaxation by progressively increasing the negative penalty with distance.
Specifically, we introduce two thresholds: an exclusion radius $R_{\mathrm{e}}$ and a saturation radius $R_{\mathrm{s}}$.
Pairs within $R_{\mathrm{e}}$ are fully suppressed, whereas pairs beyond $R_{\mathrm{s}}$ are treated as standard negatives.
For intermediate distances, the negative weight is smoothly interpolated from $0$ to $1$ to reflect the gradual decrease in semantic similarity as physical distance increases.
With curvature parameter $p \ge 1$, the weighting function is defined as:
\begin{equation}
    \omega_{\mathrm{hyb}}(d_{ij}) =
    \begin{cases}
    1 & d_{ij} = 0, \\
    0 & 0 < d_{ij} \le R_{\mathrm{e}}, \\
    \left( \frac{d_{ij} - R_{\mathrm{e}}}{R_{\mathrm{s}} - R_{\mathrm{e}}} \right)^p & R_{\mathrm{e}} < d_{ij} < R_{\mathrm{s}}, \\
    1 & d_{ij} \ge R_{\mathrm{s}}.
    \end{cases}
    \label{eq:omega_hybrid}
\end{equation}

This formulation generalizes standard InfoNCE, which can be viewed as the special case where $R_{\mathrm{e}}=R_{\mathrm{s}}=0$ or equivalently $\omega(\cdot)\equiv 1$.

\section{Experiments}
Our evaluation includes: (i) main comparisons on KITTI and KITTI-360, (ii) ablations on DIV and SC-InfoNCE, (iii) robustness to architectural choices, and (iv) qualitative visualization of cross-modal alignment.

\subsection{Evaluation Protocol}
\label{sec:eval_protocol}
\paragraph{Datasets.}
We evaluate on KITTI-360 and KITTI Odometry.
Following the common protocol in UniLoc \cite{Xia24UniLoc}, we use KITTI-360 for training and in-domain evaluation, with sequences 00, 02, 04, 06, and 07 for training, sequence 10 for validation, and sequences 03, 05, and 09 for testing.
KITTI-360 \cite{Liao23KITTI360} provides synchronized LiDAR--image pairs captured with a Velodyne HDL-64E and a front-facing perspective stereo camera, together with accurate IMU/GPS-based vehicle poses.
To assess cross-dataset generalization, we additionally evaluate on KITTI Odometry \cite{Geiger12KITTI} sequences 00, 02, 07, and 08.
Since the driving routes of KITTI and KITTI-360 do not overlap, cross-dataset evaluation measures generalization across different environments.

\paragraph{Metrics.}
We evaluate retrieval performance using Recall@K (R@K), defined as the percentage of queries for which at least one of the top-$K$ retrieved database elements falls within a spatial distance threshold $\delta$ of the ground-truth location.
Unless otherwise stated, we report R@1 under a fixed threshold of $\delta = 10\,\mathrm{m}$. To avoid trivial matches, we exclude the cross-modal paired element captured at the same frame as the query from being counted as a correct retrieval.

\paragraph{Implementation Details.}
By default, we employ DINOv1 ViT-S/16 as the backbone for both the image and LiDAR branches. Following standard practice, RGB images and DIVs are resized to $224 \times 224$ for the ViT encoders. See Appendix~\ref{app:imp} for full implementation details.

\subsection{Comparison with SOTA Methods}
\label{sec:sota_comparison}

\begin{table*}[t]
\caption{ Comparison to SOTA methods on KITTI and KITTI-360 datasets. The best is highlighted in \textbf{bold} and the second is \underline{underlined}. Results for UniLoc and Cross-PRNet are taken from the original papers.
GeoUniPR uses DIV with SALAD. Our variants differ in backbone and contrastive loss:
(v1) DINOv1+InfoNCE, (v2) DINOv1+SC-Hard, (v3) DINOv1+SC-Hybrid,
 (v4) DINOv3+SC-Hard.}
\label{tab:main_results}
\vskip 0.10in
\centering
\scriptsize
\resizebox{\textwidth}{!}{
\setlength{\tabcolsep}{3pt}
\renewcommand{\arraystretch}{1.05}
\begin{threeparttable}
\begin{tabular}{lcccccccc|cccccccc}
\toprule
\addlinespace[2pt]

& \multicolumn{8}{c|}{KITTI (Generalization)} & \multicolumn{8}{c}{KITTI-360} \\
\cmidrule(lr){2-9}\cmidrule(lr){10-17}

Method
& \multicolumn{2}{c}{2D$\to$3D}
& \multicolumn{2}{c}{3D$\to$2D}
& \multicolumn{2}{c}{2D$\to$2D}
& \multicolumn{2}{c|}{3D$\to$3D}
& \multicolumn{2}{c}{2D$\to$3D}
& \multicolumn{2}{c}{3D$\to$2D}
& \multicolumn{2}{c}{2D$\to$2D}
& \multicolumn{2}{c}{3D$\to$3D} \\
\cmidrule(lr){2-3}\cmidrule(lr){4-5}\cmidrule(lr){6-7}\cmidrule(lr){8-9}
\cmidrule(lr){10-11}\cmidrule(lr){12-13}\cmidrule(lr){14-15}\cmidrule(lr){16-17}

& R@1 & R@5
& R@1 & R@5
& R@1 & R@5
& R@1 & R@5
& R@1 & R@5
& R@1 & R@5
& R@1 & R@5
& R@1 & R@5 \\

\midrule

LiP-Loc     & 62.82 & 83.32 & 63.10 & 83.89 & 98.94 & 99.91 & 96.47 & 99.60 & 73.37 & 89.51 & 73.50 & 90.83 & 97.71 & 99.45 & 97.74 & 99.67 \\
VXP         & 23.11    & 30.68 & 17.28 & 25.52 & \textbf{99.85} & \textbf{100.00} & 97.77 & 99.82 & 33.25 & 42.17 & 27.41 & 35.88 & \textbf{99.75} & \textbf{99.96} & 98.22 & \underline{99.86} \\
\textcolor{gray}{UniLoc$^{\dagger}$}     & \textcolor{gray}{--}    & \textcolor{gray}{--} & \textcolor{gray}{--} & \textcolor{gray}{--} & \textcolor{gray}{--} & \textcolor{gray}{--} & \textcolor{gray}{--} & \textcolor{gray}{--} & \textcolor{gray}{93.50} & \textcolor{gray}{97.70} & \textcolor{gray}{94.40} & \textcolor{gray}{98.00} & \textcolor{gray}{96.50} & \textcolor{gray}{98.90} & \textcolor{gray}{97.20} & \textcolor{gray}{99.20} \\
Cross-PRNet$^*$ & 84.66    & 92.72 & -- & -- & -- & -- & -- & -- & 89.58 & 96.79 & -- & -- & -- & -- & -- & -- \\

\midrule

GeoUniPR (v1) & 94.93 & 99.14 & 93.85 & 98.65 & 99.28 & 99.96 & 98.72 & 99.89 & 95.73 & 99.24 & 95.45 & 99.18 & 98.01 & 99.70 & 97.90 & 99.65 \\
GeoUniPR (v2) & 95.90 & \underline{99.30} & 94.75 & 98.68 & \underline{99.71} & \underline{99.99} & \underline{99.29} & 99.94 & 96.97 & \underline{99.45} & 96.94 & 99.37 & 98.99 & \underline{99.92} & \underline{98.82} & 99.82 \\
GeoUniPR (v3) & \underline{96.08} & 99.26 & \underline{95.62} & \underline{99.03} & 99.67 & \underline{99.99} & 99.22 & \underline{99.95} & \underline{97.29} & \textbf{99.58} & \underline{97.03} & \underline{99.51} & \underline{99.00} & 99.87 & 98.79 & 99.85 \\
GeoUniPR (v4) & \textbf{97.20} & \textbf{99.79} & \textbf{96.97} & \textbf{99.59} & 99.65 & \textbf{100.00} & \textbf{99.38} & \textbf{99.96} & \textbf{97.41} & \textbf{99.58} & \textbf{97.49} & \textbf{99.60} & 98.85 & 99.86 & \textbf{98.91} & \textbf{99.87} \\
\addlinespace[2pt]
\bottomrule
\end{tabular}
\begin{tablenotes}
\item[$\dagger$] UniLoc results are reported under a 20\,m threshold.
\item[*] Cross-PRNet is trained on KITTI and KITTI-360 separately, and evaluated on their corresponding test datasets.
\end{tablenotes}
\end{threeparttable}
}
\vskip 0.10in
\end{table*}

\Cref{tab:main_results} summarizes cross-modal place recognition results on KITTI and KITTI-360 under a unified evaluation protocol.
We compare against representative SOTA methods, including LiP-Loc \cite{Puligilla24LIPLoc}, VXP \cite{Li25VXP}, UniLoc \cite{Xia24UniLoc}, and Cross-PRNet \cite{Meng25ContrastiveIF}.
Among them, LiP-Loc and VXP are trained under the same training/validation splits as our setting.
Note that the VXP performance reported here differs from the original paper because we evaluate it under the same cross-dataset protocol as GeoUniPR, rather than the original VXP split.
Results for UniLoc and Cross-PRNet are taken from the original papers; UniLoc reports results at a 20\,m threshold, whereas all other methods are evaluated at the stricter 10\,m threshold.
On KITTI-360, GeoUniPR attains the highest retrieval accuracy. Our best variant (v4) achieves 97.41\% and 97.49\% R@1 for 2D$\to$3D and 3D$\to$2D retrieval, respectively, distinctly outperforming other methods.
On the KITTI dataset, our method demonstrates remarkable robustness, where v4 reaches 97.20\% (2D$\to$3D) and 96.97\% (3D$\to$2D).
Moreover, all our DINOv1-based variants (v1--v3) consistently exceed the performance of existing SOTA methods across these cross-modal benchmarks.

\paragraph{Cross-dataset generalization.}
GeoUniPR demonstrates superior performance and robustness compared to baselines on both datasets.
Notably, while methods like LiP-Loc achieve reasonable in-domain results on KITTI-360, they experience a sharp performance decline ($\sim$10\% drop in R@1) when evaluated on KITTI.
This degradation likely stems from their reliance on RIV, where sensor-configuration variations introduce projection-level domain shifts that disrupt the learned cross-modal alignment.
This interpretation is further supported by the controlled LiP-Loc comparison in Appendix~\ref{app:liploc_view}, where replacing RIV with DIV improves cross-modal retrieval under the same backbone and training protocol.
In contrast, GeoUniPR maintains consistent accuracy across datasets by constructing Geometry-Consistent DIVs that establish pixel-level RGB--LiDAR correspondence and reduce sensor-dependent geometric distortions at the representation level.
This result supports our central claim that reducing modality discrepancy at the representation level yields better cross-domain robustness than compensating for projection mismatch through feature-level alignment.

\subsection{Ablation Studies}
\label{sec:ablation}
We perform controlled ablations to quantify the contribution of each component in GeoUniPR.
Unless otherwise specified, all ablations are based on GeoUniPR (v2), which adopts three-channel DIV, DINOv1 ViT-S/16, adapters, SALAD, and SC-Hard with an ambiguous-region radius $R_{\mathrm{n}}=20$.
We report cross-modal Recall@1 (R@1, \%) for both 2D$\to$3D and 3D$\to$2D on KITTI and KITTI-360.

\paragraph{Geometry-Consistent DIV.}
\label{sec:view_ablation}
\begin{table*}[t]
\centering
\begin{minipage}[t]{0.48\textwidth}
\captionof{table}{Geometry-Consistent DIV and channel ablation (R@1, \%). CH1 denotes the primary geometric channel (Range for RIV and Depth for DIV), CH2 is intensity, and CH3 is Normal Ratio. Best and second best in each column are in \textbf{bold} and \underline{underlined}.}
\label{tab:view_channels}
\vskip 0.12in
\centering
\scriptsize
\setlength{\tabcolsep}{3pt}
\renewcommand{\arraystretch}{1.05}
\resizebox{\linewidth}{!}{
\begin{tabular}{lccc cc|cc}
\toprule
\multirow{2}{*}{View} & \multicolumn{3}{c}{Input Channels} & \multicolumn{2}{c|}{KITTI} & \multicolumn{2}{c}{KITTI-360} \\
\cmidrule(lr){2-4} \cmidrule(lr){5-6} \cmidrule(lr){7-8}
& CH1 & CH2 & CH3 & 2D$\to$3D & 3D$\to$2D & 2D$\to$3D & 3D$\to$2D \\
\midrule
RIV & \checkmark & \checkmark & \checkmark
& 92.51 & 90.46 & \textbf{97.19} & \textbf{97.10} \\
\midrule
\multirow{4}{*}{DIV}
& \checkmark &          &
& 86.20 & 77.34 & 94.43 & 94.37 \\
& \checkmark & \checkmark &
& 87.54 & 82.94 & 96.42 & 96.48 \\
& \checkmark &          & \checkmark
& \underline{95.56} & \underline{94.28} & 96.83 & 96.85 \\
& \checkmark & \checkmark & \checkmark
& \textbf{95.90} & \textbf{94.75} & \underline{96.97} & \underline{96.94} \\
\bottomrule
\end{tabular}
}
\end{minipage}\hfill
\begin{minipage}[t]{0.48\textwidth}
\captionof{table}{SC-InfoNCE hyperparameter study (R@1, \%). SC-Hard sweeps the ambiguous-region radius $R_{\mathrm{n}}$. SC-Hybrid fixes $R_{\mathrm{e}}=3$, $R_{\mathrm{s}}=20$ and sweeps $p$. Best and second best in each column are in \textbf{bold} and \underline{underlined}.}
\label{tab:ga_hparam}
\vskip 0.15in
\centering
\scriptsize
\setlength{\tabcolsep}{3pt}
\renewcommand{\arraystretch}{1.05}
\resizebox{\linewidth}{!}{
\begin{tabular}{lccc|cc}
\toprule
\multirow{2}{*}{Loss} & \multirow{2}{*}{Params} & \multicolumn{2}{c|}{KITTI} & \multicolumn{2}{c}{KITTI-360} \\
\cmidrule(lr){3-4}\cmidrule(lr){5-6}
& & 2D$\to$3D & 3D$\to$2D & 2D$\to$3D & 3D$\to$2D \\
\midrule
InfoNCE & --
& 94.93 & 93.85 & 95.73 & 95.45 \\
\midrule
\multirow{4}{*}{SC-Hard}
& $R_{\mathrm{n}}=\phantom{0}1$
& 95.23 & 94.94 & 95.83 & 95.96 \\
& $R_{\mathrm{n}}=\phantom{0}3$
& 95.81 & 95.10 & 96.95 & 96.83 \\
& $R_{\mathrm{n}}=10$
& 95.74 & 95.39 & \underline{97.09} & 96.98 \\
& $R_{\mathrm{n}}=20$
& 95.90 & 94.75 & 96.97 & 96.94 \\
\midrule
\multirow{3}{*}{SC-Hybrid}
& $p=1$
& \underline{96.04} & 95.07 & 97.08 & \textbf{97.17} \\
& $p=2$
& \textbf{96.08} & \textbf{95.62} & \textbf{97.29} & 97.03 \\
& $p=3$
& 95.92 & \underline{95.54} & 97.02 & \underline{97.06} \\
\bottomrule
\end{tabular}
}
\end{minipage}
\end{table*}

We isolate the impact of Geometry-Consistent representation by comparing DIV with a strong RIV baseline.
The RIV baseline is additionally aligned to the camera FOV, ensuring that the comparison focuses on the representation itself rather than FOV differences.
As shown in \cref{tab:view_channels}, depth-only DIV achieves competitive performance by establishing camera-perspective RGB--LiDAR correspondence.
Nevertheless, perspective projection alone is insufficient for robust cross-dataset generalization, as a single depth channel provides limited local structural information.

By incorporating LiDAR-native cues, multi-channel DIV substantially improves cross-modal retrieval, with the normal-ratio channel (CH3) contributing the most significant gain.
On cross-dataset KITTI evaluation, multi-channel DIV improves over depth-only DIV by 9.70 percentage points in 2D$\to$3D R@1 (86.20\% $\to$ 95.90\%) and 17.41 percentage points in 3D$\to$2D R@1 (77.34\% $\to$ 94.75\%).
Specifically, depth provides the primary camera-aligned geometry, intensity adds reflectance cues, and the normal-ratio channel captures local surface-structure variations and discontinuities that are more comparable to image-space geometric patterns.

The full multi-channel DIV further surpasses the aligned RIV baseline on cross-dataset KITTI by +3.39\% and +4.29\% for 2D$\to$3D and 3D$\to$2D retrieval, respectively.
These results show that camera-perspective projection provides a strong geometric basis, while LiDAR-native structural enrichment is crucial for achieving robust cross-dataset CMPR.

\paragraph{SC-InfoNCE Hyperparameters.}
\label{sec:ga_hparam}

We study the sensitivity of SC-InfoNCE to its key hyperparameters under the default training protocol.
We consider two instantiations:
(i) SC-Hard, which suppresses spatially ambiguous pairs within an ambiguous region of radius $R_{\mathrm{n}}$, and
(ii) SC-Hybrid, which applies distance-aware soft reweighting controlled by the exponent $p$.
For SC-Hybrid, we fix $R_{\mathrm{e}}=3$ and $R_{\mathrm{s}}=20$ following the empirical range suggested by the SC-Hard sweep.
\Cref{tab:ga_hparam} summarizes the results.

Both SC-Hard and SC-Hybrid improve over the InfoNCE baseline across datasets and retrieval directions.
SC-Hard remains robust over a wide range of $R_{\mathrm{n}}$, while SC-Hybrid provides comparable gains with a smooth relaxation.
Guided by the robustness trend in \cref{tab:ga_hparam} and the supplementary stability analysis in Appendix~\ref{app:sc_analysis}, we use $R_{\mathrm{n}}=20$ as the default setting for SC-Hard, and set $p=2$ as the default exponent for SC-Hybrid.

\paragraph{Impact of Aggregation.}
\label{sec:Aggregation_ablation}

\begin{table*}[t]
\centering
\begin{minipage}[t]{0.48\textwidth}
\captionof{table}{Aggregation ablation (R@1, \%). Best results in each column are in \textbf{bold} and \underline{underlined}.}
\label{tab:Aggregation_ablation}
\vskip 0.15in
\centering
\scriptsize
\setlength{\tabcolsep}{3pt}
\renewcommand{\arraystretch}{1.05}
\resizebox{\linewidth}{!}{
\begin{tabular}{llcc|cc}
\toprule
\multirow{2}{*}{Loss} & \multirow{2}{*}{Aggregation} & \multicolumn{2}{c|}{KITTI} & \multicolumn{2}{c}{KITTI-360} \\
\cmidrule(lr){3-4}\cmidrule(lr){5-6}
& & 2D$\to$3D & 3D$\to$2D & 2D$\to$3D & 3D$\to$2D \\
\midrule
\multirow{3}{*}{InfoNCE}
& GeM   & 77.19 & 70.00 & 88.66 & 87.92 \\
& SPM  & 79.46 & 76.47 & 88.01 & 88.09 \\
& SALAD & \underline{94.93} & \underline{93.85} & \underline{95.73} & \underline{95.45} \\
\midrule
\multirow{3}{*}{SC-Hard}
& GeM   & 80.96 & 74.93 & 90.76 & 90.80 \\
& SPM  & 82.98 & 79.06 & 92.03 & 91.51 \\
& SALAD & \textbf{95.90} & \textbf{94.75} & \textbf{96.97} & \textbf{96.94} \\
\bottomrule
\end{tabular}
}
\end{minipage}\hfill
\begin{minipage}[t]{0.48\textwidth}
\captionof{table}{Backbone scaling study (R@1, \%). All backbones are ViT-S. Best and second best in each column are in \textbf{bold} and \underline{underlined}.}
\label{tab:backbone_ablation}
\vskip 0.15in
\centering
\scriptsize
\setlength{\tabcolsep}{3pt}
\renewcommand{\arraystretch}{1.05}
\resizebox{\linewidth}{!}{
\begin{tabular}{lcc|cc}
\toprule
\multirow{2}{*}{Backbone} & \multicolumn{2}{c|}{KITTI} & \multicolumn{2}{c}{KITTI-360} \\
\cmidrule(lr){2-3}\cmidrule(lr){4-5}
& 2D$\to$3D & 3D$\to$2D & 2D$\to$3D & 3D$\to$2D \\
\midrule
DINOv1 & 95.90 & 94.75 & 96.97 & 96.94 \\
DINOv2 & \underline{96.97} & \underline{96.74} & \underline{97.16} & \underline{97.11} \\
DINOv3 & \textbf{97.20} & \textbf{96.97} & \textbf{97.41} & \textbf{97.49} \\
\bottomrule
\end{tabular}
}
\end{minipage}
\end{table*}
We study the compatibility of SC-Hard with different aggregation heads.
\Cref{tab:Aggregation_ablation} compares GeM, SPM~\cite{Lu24CricAVPR}, and SALAD under InfoNCE and SC-Hard training.
SC-Hard consistently improves retrieval accuracy across all aggregators, showing that SC-Hard is complementary to the aggregation design.
Among the tested heads, SALAD achieves the best performance, highlighting the importance of discriminative global descriptors for cross-modal alignment.

\paragraph{Backbone Scaling.}
\label{sec:backbone_ablation}
We further evaluate scalability by replacing the backbone while keeping the remaining pipeline unchanged.
Following prior place recognition frameworks~\cite{Puligilla24LIPLoc,Jung25ImLPR}, we adopt ViT-S variants of DINOv1, DINOv2, and DINOv3.
As shown in \Cref{tab:backbone_ablation}, performance improves consistently with stronger backbones across datasets and retrieval directions, demonstrating that our Geometry-Consistent input and SC-Hard are compatible with backbone scaling.

\begin{wrapfigure}{r}{0.50\textwidth}
  \centering
  \includegraphics[width=\linewidth]{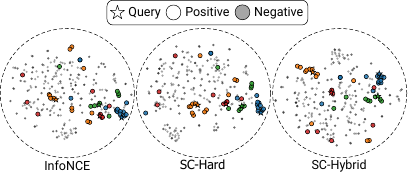}
  \caption{t-SNE visualization of 2D$\to$3D embeddings on KITTI sequence 00 under different loss functions. Query and its positive match are highlighted in color, while negatives are shown in gray.}
  \label{fig:tsne}
\end{wrapfigure}

\subsection{Qualitative Analysis}
\label{sec:qualitative}

\paragraph{t-SNE visualization.}
We qualitatively inspect the learned embedding geometry using t-SNE \cite{vanDerMaaten08tSNE} on KITTI sequence 00.
\Cref{fig:tsne} visualizes 2D$\to$3D retrieval features trained with InfoNCE, SC-Hard, and SC-Hybrid.
While most query--positive structures are similar across losses, SC-Hard and SC-Hybrid yield more compact local neighborhoods for certain queries, such as the highlighted orange one.
This indicates that SC-InfoNCE mitigates distance-induced false negatives by preserving spatially adjacent samples rather than pushing them apart as strict negatives.
SC-Hybrid further shows slightly more coherent local compactness, benefiting from its smooth distance-aware weighting.

Additional qualitative retrieval examples for both KITTI and KITTI-360 are provided in Appendix~\ref{app:qual_retrieval}.

\section{Limitations}
\label{sec:limitations}
GeoUniPR has three main limitations. First, its cross-sensor generalization is limited because DIV is directly tied to LiDAR geometry and point distribution; as a result, a model trained on HDL-64E data degrades when transferred to a different and sparser sensor such as HDL-32E. Second, multi-channel DIV construction introduces extra online cost, mainly from normal-ratio computation and densification, especially for 3D$\to$2D retrieval. Third, since DIV depends on camera-aligned projection, the method is sensitive to large extrinsic perturbations, particularly rotational noise. Additional analysis is provided in Appendix~\ref{app:limitations}.

\section{Conclusion}

We presented GeoUniPR, a unified framework for bidirectional cross-modal place recognition between RGB and LiDAR that bridges the modality gap through complementary input- and loss-level designs.
At the input level, GeoUniPR constructs Geometry-Consistent DIVs by projecting LiDAR data onto the camera plane, enforcing pixel-level alignment while incorporating surface normals as geometric priors.
At the loss level, SC-InfoNCE mitigates ambiguity from trajectory continuity and refines the cross-modal embedding space.
Experiments on KITTI and KITTI-360 demonstrate that GeoUniPR achieves SOTA performance for both 2D$\to$3D and 3D$\to$2D retrieval.
Crucially, under reliable camera--LiDAR calibration, GeoUniPR outperforms spherical projection methods in cross-dataset generalization, supporting the benefit of representation-level geometric alignment.
Future work will focus on improving deployment robustness through cross-sensor generalization, more efficient DIV construction, and calibration-robust training.

\begingroup
\small
\bibliography{references}

\begin{thebibliography}{26}
\providecommand{\natexlab}[1]{#1}
\providecommand{\url}[1]{\texttt{#1}}
\expandafter\ifx\csname urlstyle\endcsname\relax
  \providecommand{\doi}[1]{doi: #1}\else
  \providecommand{\doi}{doi: \begingroup \urlstyle{rm}\Url}\fi

\bibitem[Yin et~al.(2025)Yin, Jiao, Zhao, Xu, Huang, Choset, Scherer, and
  Han]{Yin25GPRSurvey}
Peng Yin, Jianhao Jiao, Shiqi Zhao, Lingyun Xu, Guoquan Huang, Howie Choset,
  Sebastian Scherer, and Jianda Han.
\newblock General place recognition survey: Towards real-world autonomy.
\newblock \emph{{IEEE} Transactions on Robotics}, 41:\penalty0 3019--3038,
  2025.
\newblock \doi{10.1109/TRO.2025.3550771}.

\bibitem[Cadena et~al.(2016)Cadena, Carlone, Carrillo, Latif, Scaramuzza,
  Neira, Reid, and Leonard]{Cadena16SLAMSurvey}
C{\'e}sar Cadena, Luca Carlone, Henry Carrillo, Yasir Latif, Davide Scaramuzza,
  Jos{\'e} Neira, Ian~D. Reid, and John~J. Leonard.
\newblock Past, present, and future of simultaneous localization and mapping:
  Toward the robust-perception age.
\newblock \emph{{IEEE} Transactions on Robotics}, 32\penalty0 (6):\penalty0
  1309--1332, 2016.
\newblock \doi{10.1109/TRO.2016.2624754}.

\bibitem[Newman et~al.(2002)Newman, Leonard, Tard{\'o}s, and
  Neira]{Newman02ExploreReturn}
Paul~M. Newman, John~J. Leonard, Juan~D. Tard{\'o}s, and Jos{\'e} Neira.
\newblock Explore and return: Experimental validation of real-time concurrent
  mapping and localization.
\newblock In \emph{Proceedings of the {IEEE} International Conference on
  Robotics and Automation ({ICRA})}, pages 1802--1809. {IEEE}, 2002.
\newblock \doi{10.1109/ROBOT.2002.1014803}.

\bibitem[Ratz et~al.(2020)Ratz, Dymczyk, Siegwart, and Dub{\'e}]{Ratz20OneShot}
Sebastian Ratz, Marcin Dymczyk, Roland Siegwart, and Renaud Dub{\'e}.
\newblock {OneShot} global localization: Instant {LiDAR}-visual pose
  estimation.
\newblock In \emph{Proceedings of the {IEEE} International Conference on
  Robotics and Automation ({ICRA})}, pages 5415--5421. {IEEE}, 2020.
\newblock \doi{10.1109/ICRA40945.2020.9197458}.

\bibitem[Sarlin et~al.(2019)Sarlin, Cadena, Siegwart, and
  Dymczyk]{Sarlin19CoarseToFine}
Paul-Edouard Sarlin, C{\'e}sar Cadena, Roland Siegwart, and Marcin Dymczyk.
\newblock From coarse to fine: Robust hierarchical localization at large scale.
\newblock In \emph{Proceedings of the {IEEE}/{CVF} Conference on Computer
  Vision and Pattern Recognition ({CVPR})}, pages 12716--12725, 2019.

\bibitem[Barros et~al.(2021)Barros, Pereira, Garrote, Premebida, and
  Nunes]{Barros21PRSurvey}
Tiago Barros, Ricardo Pereira, Lu{\'\i}s Garrote, Cristiano Premebida, and
  Urbano~J. Nunes.
\newblock Place recognition survey: An update on deep learning approaches.
\newblock \emph{CoRR}, abs/2106.10458, 2021.
\newblock URL \url{https://arxiv.org/abs/2106.10458}.

\bibitem[Xia et~al.(2024)Xia, Li, Li, Shi, Cao, Henriques, and
  Cremers]{Xia24UniLoc}
Yifan Xia, Ziyuan Li, Y.~J. Li, Lei Shi, H.~Cao, Joao~F. Henriques, and Daniel
  Cremers.
\newblock {UniLoc}: Towards universal place recognition using any single
  modality.
\newblock \emph{arXiv preprint}, 2024.

\bibitem[Jiao et~al.(2025)Jiao, Su, Luo, Yu, Zhou, Lu, and Chen]{Jiao25InsCMPR}
Sheng Jiao, Zhenyu Su, Luyang Luo, Haoyu Yu, Zihan Zhou, Huimin Lu, and
  Xieyuanli Chen.
\newblock {InsCMPR}: Efficient cross-modal place recognition via instance-aware
  hybrid mamba-transformer.
\newblock In \emph{Proceedings of the {IEEE} International Conference on
  Robotics and Automation ({ICRA})}, pages 2212--2218. {IEEE}, 2025.

\bibitem[Lee et~al.(2023)Lee, Song, Lim, Lee, and
  Myung]{Lee23LiDARCameraLoopConstraints}
A.~J. Lee, Seunghyeon Song, Hyungtae Lim, Wooseok Lee, and Hyun Myung.
\newblock {LiDAR}-camera loop constraints for cross-modal place recognition.
\newblock \emph{{IEEE} Robotics and Automation Letters}, 8\penalty0
  (6):\penalty0 3589--3596, 2023.

\bibitem[Puligilla et~al.(2024)Puligilla, Omama, Zaidi, Parihar, and
  Krishna]{Puligilla24LIPLoc}
S.~S. Puligilla, M.~Omama, H.~Zaidi, U.~S. Parihar, and M.~Krishna.
\newblock {LIP-Loc}: {LiDAR} image pretraining for cross-modal localization.
\newblock In \emph{Proceedings of the {IEEE}/{CVF} Winter Conference on
  Applications of Computer Vision Workshops ({WACVW})}, pages 939--948. {IEEE},
  2024.

\bibitem[Li et~al.(2025)Li, Gladkova, Xia, Wang, and Cremers]{Li25VXP}
Y.~J. Li, Maria Gladkova, Yifan Xia, Rui Wang, and Daniel Cremers.
\newblock {VXP}: Voxel-cross-pixel large-scale camera-{LiDAR} place
  recognition.
\newblock In \emph{Proceedings of the International Conference on {3D} Vision
  ({3DV})}, pages 1233--1242. {IEEE}, 2025.

\bibitem[Zheng et~al.(2023)Zheng, Li, Yu, Yu, Cao, Wang, et~al.]{Zheng23I2PRec}
Shen Zheng, Yiming Li, Zhihao Yu, Bowen Yu, Shuangyuan Cao, M.~Wang, et~al.
\newblock {I2P-Rec}: Recognizing images on large-scale point cloud maps through
  bird's eye view projections.
\newblock In \emph{Proceedings of the {IEEE}/{RSJ} International Conference on
  Intelligent Robots and Systems ({IROS})}, pages 1395--1400. {IEEE}, 2023.

\bibitem[Caron et~al.(2021)Caron, Touvron, Misra, Jegou, Mairal, Bojanowski,
  and Joulin]{Caron21DINO}
Mathilde Caron, Hugo Touvron, Ishan Misra, Herve Jegou, Julien Mairal, Piotr
  Bojanowski, and Armand Joulin.
\newblock Emerging properties in self-supervised vision transformers.
\newblock In \emph{Proceedings of the IEEE/CVF International Conference on
  Computer Vision ({ICCV})}, pages 9650--9660, 2021.

\bibitem[Oquab et~al.(2023)Oquab, Darcet, Moutakanni, Vo, Szafraniec, Khalidov,
  Fernandez, Haziza, Massa, El-Nouby, et~al.]{Oquab23DINOv2}
Maxime Oquab, Timothee Darcet, Theo Moutakanni, Huy Vo, Marcin Szafraniec,
  Vasil Khalidov, Pierre Fernandez, Daniel Haziza, Francisco Massa, Alaaeldin
  El-Nouby, et~al.
\newblock {DINO}v2: Learning robust visual features without supervision.
\newblock \emph{arXiv preprint}, 2023.

\bibitem[Simeoni et~al.(2025)Simeoni, Vo, Seitzer, Baldassarre, Oquab, Jose,
  et~al.]{Simeoni25DINOv3}
Oriane Simeoni, Huy~V. Vo, Maximilian Seitzer, Federico Baldassarre, Maxime
  Oquab, Cesar Jose, et~al.
\newblock {DINO}v3.
\newblock \emph{arXiv preprint}, 2025.

\bibitem[Lu et~al.(2024)Lu, Lan, Zhang, Jiang, Wang, and Yuan]{Lu24CricAVPR}
Feng Lu, Xiaoyan Lan, Lei Zhang, Dong Jiang, Yao Wang, and Chun Yuan.
\newblock {CricAVPR}: Cross-image correlation-aware representation learning for
  visual place recognition.
\newblock In \emph{Proceedings of the IEEE/CVF Conference on Computer Vision
  and Pattern Recognition ({CVPR})}, pages 16772--16782, 2024.

\bibitem[Lu et~al.(2025)Lu, Jin, Lan, Zhang, Liu, Wang, and
  Yuan]{Lu25SelaVPRpp}
Feng Lu, Tian Jin, Xiaoyan Lan, Lei Zhang, Yuting Liu, Yao Wang, and Chun Yuan.
\newblock {SelaVPR}++: Towards seamless adaptation of foundation models for
  efficient place recognition.
\newblock \emph{arXiv preprint}, 2025.

\bibitem[Radford et~al.(2021)Radford, Kim, Hallacy, Ramesh, Goh, Agarwal,
  Sastry, Askell, Mishkin, Clark, Krueger, and Sutskever]{Radford21CLIP}
Alec Radford, Jong~Wook Kim, Chris Hallacy, Aditya Ramesh, Gabriel Goh,
  Sandhini Agarwal, Girish Sastry, Amanda Askell, Pamela Mishkin, Jack Clark,
  Gretchen Krueger, and Ilya Sutskever.
\newblock Learning transferable visual models from natural language
  supervision.
\newblock In \emph{Proceedings of the 38th International Conference on Machine
  Learning ({ICML})}, pages 8748--8763. {PMLR}, 2021.

\bibitem[Cattaneo et~al.(2020)Cattaneo, Vaghi, Fontana, Ballardini, and
  Sorrenti]{Cattaneo20SharedEmbedding}
Davide Cattaneo, Marco Vaghi, Simone Fontana, Alessandro~L. Ballardini, and
  Domenico~G. Sorrenti.
\newblock Global visual localization in {LiDAR}-maps through shared {2D}-{3D}
  embedding space.
\newblock In \emph{Proceedings of the {IEEE} International Conference on
  Robotics and Automation ({ICRA})}, pages 4365--4371. {IEEE}, 2020.

\bibitem[Meng et~al.(2025)Meng, Wang, Xu, and Chau]{Meng25ContrastiveIF}
S.~Meng, Y.~Wang, H.~Xu, and Lap-Pui Chau.
\newblock Contrastive learning-based place descriptor representation for
  cross-modality place recognition.
\newblock \emph{Information Fusion}, page 103351, 2025.

\bibitem[Izquierdo and Civera(2024)]{Izquierdo24SALAD}
Sergio Izquierdo and Javier Civera.
\newblock Optimal transport aggregation for visual place recognition.
\newblock In \emph{Proceedings of the {IEEE}/{CVF} Conference on Computer
  Vision and Pattern Recognition ({CVPR})}, pages 17658--17668, 2024.

\bibitem[Jung et~al.(2025)Jung, Fu, Fallon, and Kim]{Jung25ImLPR}
Minwoo Jung, Lanke Frank~Tarimo Fu, Maurice Fallon, and Ayoung Kim.
\newblock {ImLPR}: Image-based {LiDAR} place recognition using vision
  foundation models.
\newblock In \emph{Proceedings of the Conference on Robot Learning ({CoRL})},
  Seoul, September 2025.

\bibitem[Ku et~al.(2018)Ku, Harakeh, and Waslander]{Ku18DepthCompletionCPU}
Jason Ku, Ali Harakeh, and Steven~L. Waslander.
\newblock In defense of classical image processing: Fast depth completion on
  the {CPU}.
\newblock In \emph{Proceedings of the 15th Conference on Computer and Robot
  Vision ({CRV})}, pages 16--22. {IEEE}, 2018.

\bibitem[Liao et~al.(2023)Liao, Xie, and Geiger]{Liao23KITTI360}
Yiyi Liao, Jun Xie, and Andreas Geiger.
\newblock {KITTI}-360: A novel dataset and benchmarks for urban scene
  understanding in {2D} and {3D}.
\newblock \emph{{IEEE} Transactions on Pattern Analysis and Machine
  Intelligence}, 45\penalty0 (3):\penalty0 3292--3310, 2023.
\newblock \doi{10.1109/TPAMI.2022.3179507}.

\bibitem[Geiger et~al.(2012)Geiger, Lenz, and Urtasun]{Geiger12KITTI}
Andreas Geiger, Philip Lenz, and Raquel Urtasun.
\newblock Are we ready for autonomous driving? the {KITTI} vision benchmark
  suite.
\newblock In \emph{Proceedings of the {IEEE} Conference on Computer Vision and
  Pattern Recognition ({CVPR})}, pages 3354--3361. {IEEE}, 2012.

\bibitem[van~der Maaten and Hinton(2008)]{vanDerMaaten08tSNE}
Laurens van~der Maaten and Geoffrey Hinton.
\newblock Visualizing data using {t-SNE}.
\newblock \emph{Journal of Machine Learning Research}, 9:\penalty0 2579--2605,
  2008.

\end{thebibliography}
\bibliographystyle{unsrtnat}
\endgroup

\newpage
\appendix
\onecolumn

\paragraph{Appendix Overview.}
This appendix is organized as follows:
\begin{itemize}
  \item Appendix~\ref{app:imp} details the implementation setup, including the network architecture, DIV construction pipeline, and training configuration.
  \item Appendix~\ref{app:sc_analysis} provides supplementary training-dynamics and run-to-run stability analyses for SC-InfoNCE.
  \item Appendix~\ref{app:liploc_view} provides an additional controlled comparison between RIV and DIV within LiP-Loc to further isolate the effect of the LiDAR view representation.
  \item Appendix~\ref{app:limitations} presents supplementary analyses of robustness to calibration perturbations and practical efficiency during training and inference.
  \item Appendix~\ref{app:qual_retrieval} shows qualitative retrieval examples on KITTI and KITTI-360.
\end{itemize}

\section{Implementation Details}
\label{app:imp}

\paragraph{Network Architecture and Variants.}
By default, we employ DINOv1 ViT-S/16 as the backbone for both RGB and LiDAR modalities, utilizing separate parameters to accommodate domain-specific distributions. We adopt a parameter-efficient fine-tuning (PEFT) protocol, where only the last two transformer blocks are fine-tuned, and lightweight MultiConv adapters are attached in parallel to every third transformer block, keeping the remaining blocks frozen. Feature aggregation is performed using a SALAD  head to extract global descriptors. Specifically, the SALAD head is configured with $M=64$ clusters, each possessing a feature dimension of $128$, while the global [CLS] token is projected to a separate $256$-dimensional embedding. To assess the scalability of our approach, we also benchmark against DINOv2 Vit-S/14 and DINOv3 Vit-S/16 backbones. Crucially, when transitioning between these backbones, the downstream architecture---including the placement of adapters and the specific parameterization of the SALAD head (i.e., cluster count and dimensions)---remains \textit{strictly identical}, ensuring that performance differences are solely attributable to the visual foundation models themselves.

\paragraph{DIV Generation and Preprocessing.}
The LiDAR input is projected into a Depth Image View (DIV) comprising three physical channels: metric depth, intensity, and normal ratio. Specifically, the normal ratio is derived from the logarithmic ratio of the largest to smallest singular values, $\ln(\sigma_1/\sigma_3)$, calculated via Singular Value Decomposition (SVD) on the covariance matrix of $k$-nearest neighbors ($k=15$). To resolve projection sparsity, these channels are processed using the training-free IP-Basic densification method \cite{Ku18DepthCompletionCPU} in its default ``fast'' variant. To ensure cross-modal geometric alignment, both RGB and DIV inputs are cropped to remove the upper non-overlapping region and resized to a canonical resolution of $224 \times 224$.

\paragraph{Training Configuration.}
All models are trained for 50 epochs using a symmetric contrastive loss with a temperature $\tau=0.07$ and a batch size of 64. Optimization is performed using AdamW (weight decay $1\times10^{-3}$) with an initial learning rate of $1\times10^{-4}$, controlled by a ReduceLROnPlateau scheduler. All experiments are implemented in PyTorch and conducted on a single NVIDIA GeForce RTX 3090 GPU.

\section{Additional Training Dynamics and Stability Analysis}
\label{app:sc_analysis}

\paragraph{Effectiveness of SC-InfoNCE.}
As shown in \Cref{tab:main_results}, SC-InfoNCE (GeoUniPR (v2-v3)) consistently improve cross-modal retrieval performance by effectively suppressing distance-induced false negatives.
For example, on KITTI-360, SC-Hybrid (GeoUniPR (v3)) improves 2D$\to$3D R@1 from 95.73\% to 97.29\%, yielding an absolute gain of 1.56 percentage points, with similar gains for 3D$\to$2D.

\begin{figure}[t]
  \centering
  \includegraphics[width=0.62\linewidth]{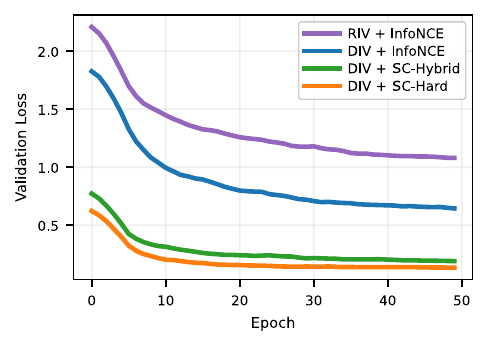}
  \caption{Validation loss curves across various combinations of image-view representations and contrastive loss formulations; DIV reduces the loss difficulty compared to RIV, and SC-InfoNCE further improves convergence by suppressing distance-induced false negatives.}
  \label{fig:loss_view_loss_ablation}
\end{figure}

\Cref{fig:loss_view_loss_ablation} shows that SC-Hard/SC-Hybrid converge faster and achieve substantially lower validation loss than InfoNCE baselines, validating the effectiveness of combining explicit alignment DIV with Spatially-Consistent supervision.

\begin{figure}[t]
  \centering
  \includegraphics[width=0.92\linewidth]{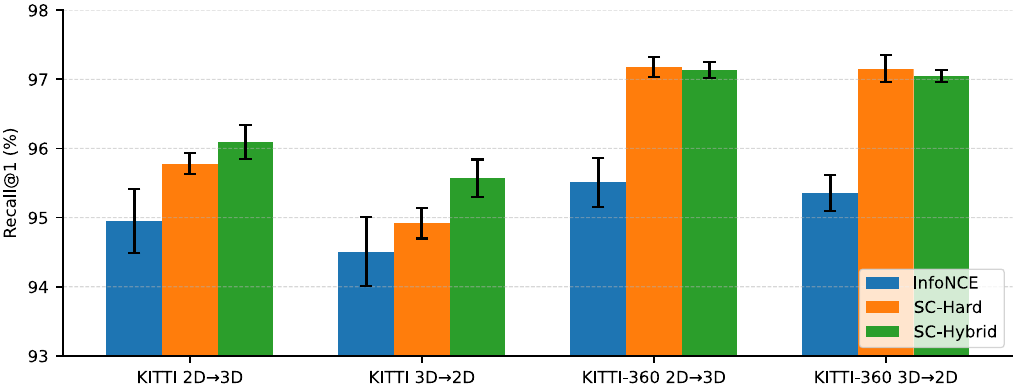}
  \caption{Cross-modal R@1 (mean$\pm$std over five runs) on KITTI and KITTI-360; SC-Hard yields lower variance, while SC-Hybrid attains higher mean accuracy under cross-dataset evaluation.}
  \label{fig:loss_errorbar}
\end{figure}

\paragraph{Stability Across Runs.}
\Cref{fig:loss_errorbar} reports cross-modal R@1 over five independent runs using mean$\pm$std.

Across both datasets and retrieval directions, SC-Hard consistently shows the smallest standard deviation, indicating more stable performance under repeated training.
Meanwhile, SC-Hybrid achieves a higher mean accuracy on KITTI (cross-dataset evaluation) for both 2D$\to$3D and 3D$\to$2D retrieval, while remaining competitive on KITTI-360.
A possible interpretation is that the soft distance-aware weighting provides a less abrupt treatment of spatially nearby samples than hard masking, which may lead to representations that generalize better across dataset shifts.

\section{Single-Channel RIV vs.\ DIV in LiP-Loc}
\label{app:liploc_view}
\begin{figure}[htbp]
    \centering
    \includegraphics[width=0.98\linewidth]{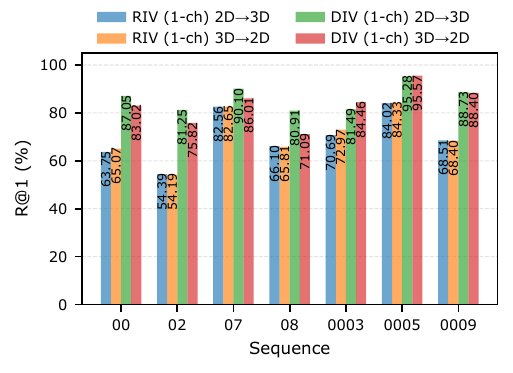}
    \caption{Cross-modal retrieval accuracy (R@1, \%) of LiP-Loc under two \emph{single-channel} LiDAR input parameterizations.
    We compare 1-channel RIV (range-only spherical RIV) and 1-channel DIV (depth-only perspective DIV) on both retrieval directions (2D$\to$3D and 3D$\to$2D).
    The x-axis lists evaluation sequences from two datasets: KITTI (00/02/07/08) and KITTI-360 (0003/0005/0009).}
    \label{fig:liploc_riv_vs_div_singlech}
\end{figure}

\Cref{fig:liploc_riv_vs_div_singlech} is designed to isolate the effect of the LiDAR view representation under a strong, closely related CMPR baseline. We adopt LiP-Loc, which performs full fine-tuning of a pretrained ViT backbone for cross-modal place recognition, and keep the architecture and training protocol fixed. Under the same single-channel LiDAR input budget, we only replace the LiDAR view from range-only RIV to depth-only DIV. Across KITTI and KITTI-360, this substitution yields higher cross-modal R@1 for both 2D$\to$3D and 3D$\to$2D, indicating that a camera-plane depth view provides a more compatible geometric signal for RGB--LiDAR matching than a spherical range view under the same fine-tuning setup.

\section{Additional Robustness and Efficiency Analysis}
\label{app:limitations}

This appendix complements the main paper with additional analysis of GeoUniPR(v2) from two practical perspectives: robustness to imperfect sensor calibration and computational efficiency during training and inference.

\subsection{Robustness to Calibration Perturbations}
\label{app:calibration_noise}

Robustness to calibration errors is important for real-world deployment because the proposed depth image view (DIV) is generated by projecting LiDAR points into the camera frame. To evaluate this dependency, we perturb the camera extrinsics \emph{only at inference time}, without introducing perturbation-aware augmentation during training. For each preset interval, the perturbation is randomly sampled within the specified range (e.g., $[-1^\circ,1^\circ]$), rather than evaluated only at the boundary values.

\begin{table}[t]
\caption{Robustness of GeoUniPR(v2) to inference-time camera extrinsic perturbations. Perturbations are randomly sampled within each preset interval rather than evaluated only at the boundary values. No perturbation-aware training is used in this experiment.}
\label{tab:calibration_noise}
\centering
\setlength{\tabcolsep}{3pt}
\renewcommand{\arraystretch}{1.05}
\begin{tabular}{cccccc}
\toprule
\multirow{2}{*}{\textbf{Rot ($^\circ$)}} & \multirow{2}{*}{\textbf{Trans (m)}} & \multicolumn{2}{c}{\textbf{KITTI-00 R@1 (\%)}} & \multicolumn{2}{c}{\textbf{KITTI-360-0003 R@1 (\%)}} \\
\cmidrule(lr){3-4} \cmidrule(lr){5-6}
& & \textbf{2D$\to$3D} & \textbf{3D$\to$2D} & \textbf{2D$\to$3D} & \textbf{3D$\to$2D} \\
\midrule
\textbf{0}         & \textbf{0}                & \textbf{97.62} & \textbf{97.38} & \textbf{88.42} & \textbf{86.24} \\
$[-1,1]$           & $0$                       & 96.28          & 96.50          & 84.65          & 84.85 \\
$[-2,2]$           & $0$                       & 93.75          & 93.64          & 79.50          & 75.64 \\
$[-5,5]$           & $0$                       & 78.07          & 64.81          & 59.21          & 41.98 \\
$0$                & $[-0.05,0.05]$           & 97.56          & 97.20          & 87.62          & 86.34 \\
$0$                & $[-0.1,0.1]$             & 97.25          & 97.14          & 85.74          & 85.54 \\
$0$                & $[-0.2,0.2]$             & 96.26          & 96.37          & 82.87          & 82.87 \\
$[-1,1]$           & $[-0.05,0.05]$           & 96.52          & 96.39          & 85.05          & 83.96 \\
$[-2,2]$           & $[-0.1,0.1]$             & 93.42          & 93.02          & 79.31          & 75.25 \\
$[-5,5]$           & $[-0.2,0.2]$             & 77.36          & 62.45          & 60.30          & 41.19 \\
\bottomrule
\end{tabular}
\end{table}

Table~\ref{tab:calibration_noise} shows that GeoUniPR(v2) remains stable under mild projection errors. On KITTI-00, the $R@1$ score stays above $96\%$ under $\pm1^\circ$ rotation or $\pm0.2$\,m translation in both retrieval directions. On KITTI-360-0003, the score remains above $84\%$ under $\pm1^\circ$ rotation or $\pm0.1$\,m translation. As the perturbation magnitude increases, the performance degrades progressively. Translation noise is consistently less harmful than rotational noise, whereas larger rotational perturbations (e.g., $\pm5^\circ$) cause a much clearer drop because they more directly disrupt perspective alignment during DIV construction. Overall, these results indicate that GeoUniPR(v2) is reasonably tolerant to mild-to-moderate calibration errors, while larger miscalibration mainly affects projection quality.

\subsection{Training Cost and Parameter Efficiency}
\label{app:training_efficiency}

\begin{table}[t]
\caption{Training cost comparison with LiP-Loc. GeoUniPR(v2) uses PEFT and therefore optimizes substantially fewer parameters despite a slightly larger total model size.}
\label{tab:training_efficiency}
\centering
\setlength{\tabcolsep}{4pt}
\renewcommand{\arraystretch}{1.05}
\begin{tabular}{lccc}
\toprule
\textbf{Method} & \textbf{Total Params (M)} & \textbf{Trainable Params (M)} & \textbf{Training Time (min/epoch)} \\
\midrule
LiP-Loc         & \textbf{42} & 42           & 4.36 \\
GeoUniPR (v2)   & 48          & \textbf{14}  & \textbf{2.56} \\
\bottomrule
\end{tabular}
\end{table}

We compare the training cost of GeoUniPR(v2) against LiP-Loc. As shown in Table~\ref{tab:training_efficiency}, GeoUniPR(v2) has a slightly larger total parameter count, but requires substantially fewer trainable parameters and less training time per epoch. This behavior is consistent with the parameter-efficient fine-tuning (PEFT) design adopted in our framework, which reduces optimization cost relative to full backbone fine-tuning. In addition, the DIV inputs used during training can be precomputed offline, further improving practical training efficiency.

\subsection{Inference Latency and Cost Breakdown}
\label{app:inference_latency}

\begin{table}[t]
\caption{Inference latency comparison between LiP-Loc and GeoUniPR(v2). GeoUniPR(v2) is competitive in 2D$\to$3D retrieval, while the 3D$\to$2D direction is slower because DIV generation must be executed online.}
\label{tab:inference_latency}
\centering
\setlength{\tabcolsep}{3pt}
\renewcommand{\arraystretch}{1.05}
\begin{tabular}{lcccc}
\toprule
\multirow{2}{*}{\textbf{Method}} & \multicolumn{2}{c}{\textbf{KITTI-00 Latency (ms)}} & \multicolumn{2}{c}{\textbf{KITTI-360-0003 Latency (ms)}} \\
\cmidrule(lr){2-3} \cmidrule(lr){4-5}
& \textbf{2D$\to$3D} & \textbf{3D$\to$2D} & \textbf{2D$\to$3D} & \textbf{3D$\to$2D} \\
\midrule
LiP-Loc        & \textbf{13.93} & \textbf{21.09} & \textbf{14.38} & \textbf{21.23} \\
GeoUniPR (v2)  & 14.13          & 81.22          & 15.15          & 86.89 \\
\bottomrule
\end{tabular}
\end{table}

Table~\ref{tab:inference_latency} compares the inference latency of LiP-Loc and GeoUniPR(v2). In the 2D$\to$3D direction, GeoUniPR(v2) is comparable to LiP-Loc on both datasets. In contrast, in the 3D$\to$2D direction, GeoUniPR(v2) is noticeably slower because the multi-channel DIV must be constructed online for each LiDAR query.

The main overhead comes from two preprocessing stages: normal-ratio computation and multi-channel densification. The average normal-ratio computation takes $47.39$\,ms on KITTI and $54.48$\,ms on KITTI-360, corresponding to an overall average of about $50.47$\,ms. The densification step adds another $21.03$\,ms on average. This breakdown explains why GeoUniPR(v2) remains efficient in 2D$\to$3D retrieval, where database descriptors can be precomputed offline, but is less efficient in 3D$\to$2D retrieval, where online DIV generation becomes part of the critical path.

Overall, these supplementary results show that GeoUniPR(v2) remains stable under mild calibration noise and achieves favorable training efficiency through PEFT, while its main practical bottleneck lies in the online construction of multi-channel DIV for 3D$\to$2D retrieval.

\section{Qualitative Retrieval Examples}
\label{app:qual_retrieval}

We visualize top-$3$ cross-modal retrieval results of GeoUniPR on representative frames from KITTI (sequences 00/02/07/08) and KITTI-360 (sequences 0003/0005/0009). For each sequence, we report both directions: 2D$\!\to\!$3D (RGB query retrieving LiDAR database) and 3D$\!\to\!$2D (LiDAR query retrieving RGB database). Green borders denote correct matches under the same evaluation tolerance used in the main paper, while red borders indicate incorrect matches.

\begin{figure}[t]
  \centering
  \includegraphics[width=0.98\linewidth]{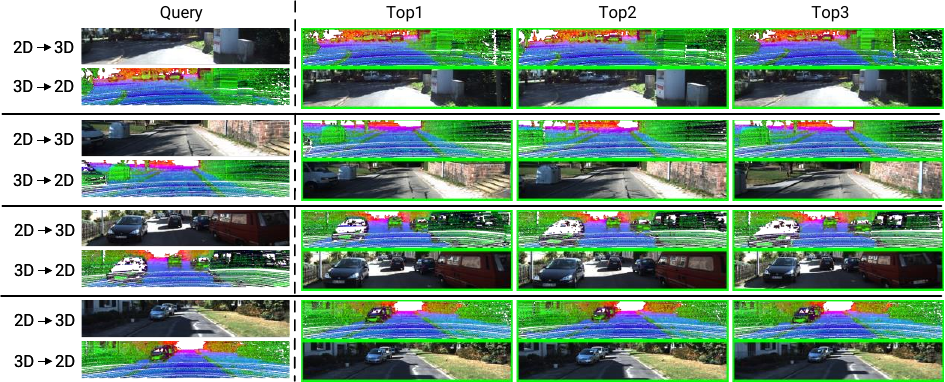}
  \caption{\textbf{KITTI qualitative top-$3$ retrievals.}
  Each \emph{pair of rows} corresponds to one sequence frame: the first row is 2D$\!\to\!$3D and the second row is 3D$\!\to\!$2D.
  From top to bottom, the four blocks correspond to KITTI sequences 00, 02, 07, and 08.
  Each row shows the query (left) and the top-3 retrieved candidates (right).
  Green borders indicate correct matches.}
  \label{fig:kitti_top3_ok}
\end{figure}

\begin{figure}[t]
  \centering
  \includegraphics[width=0.98\linewidth]{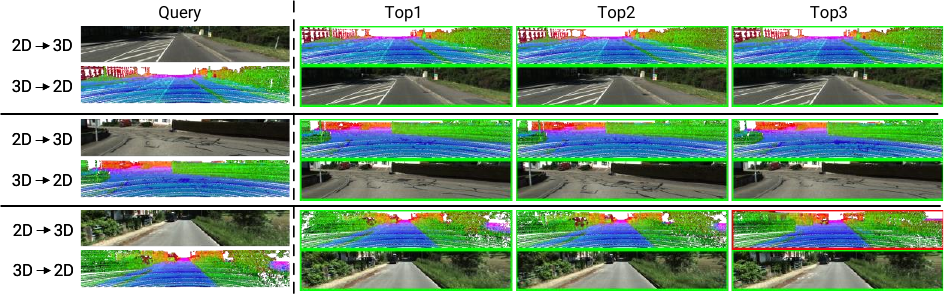}
  \caption{\textbf{KITTI-360 qualitative top-$3$ retrievals.}
  Each \emph{pair of rows} corresponds to one sequence frame: the first row is 2D$\!\to\!$3D and the second row is 3D$\!\to\!$2D.
  From top to bottom, the three blocks correspond to KITTI-360 sequences 0003, 0005, and 0009.
  Each row shows the query (left) and the top-3 retrieved candidates (right).
  Green borders denote correct matches, while red borders indicate incorrect matches.}
  \label{fig:kitti360_top3_ok}
\end{figure}

\paragraph{Correct top-$1$ retrievals.}
\cref{fig:kitti_top3_ok,fig:kitti360_top3_ok} show examples where the top-$1$ retrieval is correct.
Across both datasets and multiple sequences, GeoUniPR retrieves RGB/LiDAR candidates that remain visually and geometrically consistent with the query, and the behavior is generally stable in both directions (2D$\!\to\!$3D and 3D$\!\to\!$2D).
Qualitatively, the retrieved matches preserve salient layout cues (e.g., road geometry, curb/building boundaries, and nearby structural arrangement), suggesting that the learned embedding emphasizes transferable geometric structure rather than appearance-only similarity.

\begin{figure}[t]
  \centering
  \includegraphics[width=0.98\linewidth]{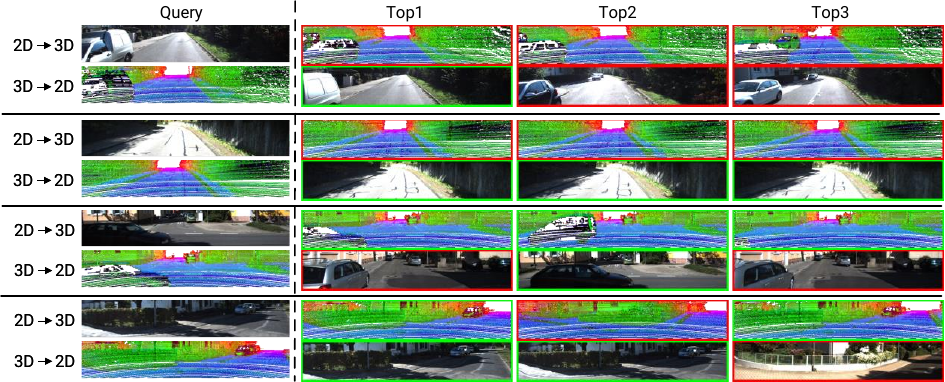}
  \caption{\textbf{KITTI qualitative top-$3$ retrievals with errors.}
  Layout follows \cref{fig:kitti_top3_ok}: for each sequence frame, the first row is 2D$\!\to\!$3D and the second row is 3D$\!\to\!$2D; blocks from top to bottom correspond to sequences 00, 02, 07, and 08.
  Green borders denote correct matches, while red borders indicate incorrect matches.}
  \label{fig:kitti_top3_neg}
\end{figure}

\begin{figure}[t]
  \centering
  \includegraphics[width=0.98\linewidth]{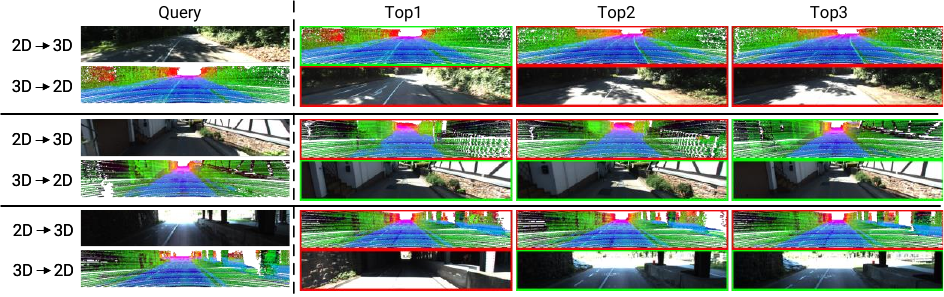}
  \caption{\textbf{KITTI-360 qualitative top-$3$ retrievals with errors.}
  Layout follows \cref{fig:kitti360_top3_ok}: for each sequence frame, the first row is 2D$\!\to\!$3D and the second row is 3D$\!\to\!$2D; blocks from top to bottom correspond to sequences 0003, 0005, and 0009.
  Green borders denote correct matches, while red borders indicate incorrect matches.}
  \label{fig:kitti360_top3_neg}
\end{figure}

\paragraph{Failure cases and characteristic confusions.}
\cref{fig:kitti_top3_neg,fig:kitti360_top3_neg} present cases where the top-$1$ or some top-$3$ candidates are incorrect. A recurrent pattern is directional asymmetry: in several frames, 2D$\!\to\!$3D is less reliable when the RGB observation lacks distinctive landmarks (e.g., weak building presence or repetitive roadside structure), while the corresponding 3D$\!\to\!$2D retrieval remains correct (and vice versa in some cases). We interpret this as modality-dependent ambiguity: appearance cues may be less discriminative in visually repetitive regions, whereas geometric cues can still constrain the match, or conversely, sparse LiDAR structure can be less informative in certain configurations. Importantly, even incorrect matches typically share highly similar global structure with the query (e.g., comparable road shape and boundary layout), indicating that most failures arise from confusing near-duplicate places rather than arbitrary mismatches.

\clearpage

\end{document}